\pdfoutput=1

\documentclass[11pt]{article}

\usepackage{ACL2023}

\usepackage{times}
\usepackage{latexsym}
\usepackage{hyperref}
\usepackage{graphicx}
\usepackage{amsmath}
\usepackage{natbib}
\usepackage[utf8]{inputenc}
\usepackage{xcolor}
\usepackage{xurl}
\usepackage{listings}
\usepackage[T1]{fontenc}

\usepackage[utf8]{inputenc}

\usepackage{microtype}

\usepackage{booktabs}

\usepackage{inconsolata}

\title{\textsc{Paramanu-Ganita}: Can Small Math Language Models Rival with Large Language Models on Mathematical Reasoning?}

\author{
    Mitodru Niyogi \\
	  Gyan AI Research \thanks{Work done at Gyan AI Research} \\
      \& \\
      CNRS, LIG, Grenoble INP, Univ. Grenoble Alpes, \\
	  Grenoble, France \\
	  {\tt mitodru.niyogi@grenoble-inp.fr}
	  \And
	Arnab Bhattacharya \\
  	Dept. of Computer Science \& Engineering, \\
  	Indian Institute of Technology Kanpur, \\
  	Kanpur, India \\
    {\tt arnabb@cse.iitk.ac.in}}

\begin{document}
\maketitle
\begin{abstract}
In this paper, we study whether domain specific pretraining of small generative language models (SLM) from scratch with domain specialized tokenizer and Chain-of-Thought (CoT) instruction fine-tuning results in competitive performance on mathematical reasoning compared to LLMs? Secondly, whether this approach is environmentally sustainable, highly cost efficient? To address these research questions, we present \textsc{Paramanu-Ganita}, a 208 million-parameter novel decoder-only Auto Regressive SLM on \emph{mathematics}. We performed pretraining from scratch on 31.5 billion tokens for 170 A100 hours using a context size of 4096 on a mixed mathematical corpus consisting of web pages, source code, textbooks, CoT templatised StackOverflow QA pairs, and mathematical lecture notes in \LaTeX{} curated by us. We also trained a math and code specialised BPE tokenizer. We proposed and performed CoT instruction fine-tuning of Paramanu-Ganita on the MetaMathQA dataset. Our model Paramanu-Ganita, despite being 34 times smaller than the 7B LLMs, outperforms generalist LLMs by approximately $30\%$ points, and even math-specialised LLMs by $3$-$23\%$ points in GSM8K test accuracy metric. On MATH benchmark, Paramanu-Ganita outperformed the various models by $6$-$8\%$ points. On benchmarks like LogiQA, MMLU (high school, college level), and competitive exams level, AGIEVAL (AQuA-RAT, SAT-Math), Paramanu-Ganita outperformed others by 1-4\%. Our model is available at \href{https://huggingface.co/gyanai/paramanu-ganita-208M-hf}{https://huggingface.co/gyanai/paramanu-ganita-208M-hf}.

\end{abstract}

\section{Introduction}

Pretrained Large Language Models (LLMs) such as LLaMa \citep{touvron2023llama}, LLaMa-2, \citep{touvron2023llama2}, PaLM \cite{JMLR:v24:22-1144}, Falcon \citep{almazrouei2023falcon}, Code LlaMa \citep{rozière2024code}, MPT \citep{databricksIntroducingMPT7B},
etc. have demonstrated multi-dimensional abilities, such as in open-ended dialogue or instruction following capabilities \citep{ouyang2022training}. Being  typically generalist language models balancing the performance across the entire distribution of natural language tasks. However, these generalist models are humongous in size and requires millions of dollars to train aside from high engineering inference cost involved. 
Traditionally, to optimize performance within specific domains such as finance \citep{wu2023bloomberggpt}, medicine \citep{singhal2023expertlevel}, etc., these models have been continually trained on domain specific data. However, domain specific continual pretraining of LLMs are also very expensive as a lot of computation and inference costs are involved along with high requirement of GPUs. For example, to improve the mathematical reasoning capabilities of LLMs, LLEMMA 7B \citep{azerbayev2024llemma} was trained on 256 A100 40GB GPUs for roughly 23,000 A100 training hours, which is extremely expensive.

We consider here that a language model is \textit{large} (LLM) if it contains more than 1B parameters and \textit{small} if it contains less than 300M parameters (SLM)\footnote{We refer to models containing more than 300M parameters and less than 1B parameters as \textit{Medium} Language Models. We however do not consider them in this study.}. In this paper, we search for a alternative approach to continual pretraining of LLMs for improving mathematical reasoning of LLMs like LLEMMA and cost-effective training and inference method. In particular, we try to answer the two following research questions.

\textbf{RQ1}: Is domain specific pretraining from scratch of small generative language model with domain specialised tokenizer and Chain-of-Thought (CoT) instruction fine-tuning results in competitive performance on mathematical reasoning compared to LLMs which are trained on trillion of tokens and humongous in size on the assumption that ``Larger models trained on trillion tokens can only reason'' parameters?

\textbf{RQ2:} Is domain specific pretraining from scratch of SLM is environmentally sustainable, highly cost efficient for both training and inference? 


To answer these questions, instead of following the domain adaptation method of LLMs for better mathematical reasoning, we focused on \emph{pretraining from scratch} a generative mathematical language model using only a high quality mathematical corpus curated by us. This avoids requiring immense compute power, high engineering maneuver and techniques to load LLMs in memory, and mostly high cost of training, and the misalignment of domain specialised tokenizers and embeddings with the existing embeddings of LLMs via continual pretraining with vocabulary expansion of the existing LLMs tokenizers.
We trained a powerful mathematical SLM from scratch which required \emph{only} 146 hours of A100 training and additional 14 hours for Chain-of-Thought (CoT) instruction fine-tuning.
We call our model \textsc{Paramanu-Ganita}\footnote{Paramanu means ``atom'' while Ganita is ``mathematics''}. 
Our model is based on the Transformer Decoder architecture \citep{10.5555/3295222.3295349}. We have trained an auto-regressive model from scratch at a context size of 4096 on a single NVidia A100-PCIE-40GB GPU. Our models are small in size, having only 208 million parameters. Hence, our models are very fast in inference without requiring any quantization of weights, and our mathematics model inference can be run on CPU without need of GPU.

To test the mathematical problem solving ability of our SLM, Paramanu-Ganita, we evaluated and compared with generalist LLMs, code LLM, and math specialized LLMs across variety of grade level difficulty benchmarks including both discriminative multiple-choice math benchmarks across competitive exams (SAT, GRE, GMAT), graduate level, high school and grade level level math questions. We also tested our model on a logical reasoning benchmark (LogiQA). Table ~\ref{tab:gsm8k-math} and Table ~\ref{tab:mcq-math} show the comparison of Paramanu-Ganita and LLMs on various mathematical benchmarks.
Although SLM, Paramanu-Ganita, still outperformed math specialised LLM like LLEMMA 7B on GSM8K \citep{cobbe2021training} benchmark despite being 35 times smaller in size. On the memory requirements, the LLEMMA 7B checkpoint size is 13.5\,GB whereas our model's checkpoint size is 2.5\,GB and less than 1\,GB in binary format (.bin).
We found that our approach is highly cost efficient as we only spent on total 170 A100 hours including both pretraining from scratch and CoT fine-tuning, making our approach to be highly cost efficient, very competitive performance wrt LLMs, and least carbon footprint compared to LLMs or even math domain specialized LLM like LLEMMA which took 23,000 A100 hours for continual pretraining of Llama 2 and yet its performance (36.40\%) is lower than our model, Paramanu-Ganita (36.77\%) on GSM8K. Therefore, with our novel approach, we cut down the training cost by 135 times without compromising the performance of the model on mathematical benchmarks compared to both generalist and math specialized LLMs.


Our main contributions in this work are as follows:

\begin{enumerate}

    \item We have curated a pretraining corpus for mathematics with high quality mathematical text from various public sources and in-house university lecture notes in \LaTeX{}, textbooks, web crawled mathematical text, mathematical source code from various programming languages (AlgebraStack), and Chain-of-Thought (CoT) \citep{wei2023chainofthought} templatised mathematical question answers pairs from forums like StackExchange. A part of our dataset is available at \href{https://huggingface.co/datasets/gyanai/ganita}{https://huggingface.co/datasets/gyanai/ganita}
    
    \item We have developed a specialised tokenizer from scratch for mathematics domain and code.

    \item We build a 208M parameter, decoder-only mathematical SLM, pretrained from scratch on the above corpus for 31.5 billion tokens at a context size of 4096. We have also performed Chain-of-Thought (CoT) instruction fine-tuning of our model on MetaMathQA \citep{yu2024metamath} dataset.
    
    

    \item Our model, Paramanu-Ganita 208M, outperformed LLaMa-1 (33B, 13B, 7B), LLaMa-2 (7B, 13B), Falcon (40B, 7B), 
    PaLM (62B, 8B), MPT (30B, 7B), Vicuna 13B, 
    and math-specialised LLMs like Minerva 8B,
    LLEMMA-7B on GSM8K, MATH, AGIEVAL-AQuA-RAT benchmarks despite being smaller by multiple orders of magnitude in size.
    
\end{enumerate}

The remainder of the paper is organized as follows: Section~\ref{sec:related-work} describes the related work, Section~\ref{sec:data} presents the data used for pretraining, and Sections~\ref{sec:model} and \ref{sec:training} the model and training procedure retained. The evaluation of our model and analysis are given in Section~\ref{sec:eval}, and Section~\ref{sec:analysis}; Section~\ref{sec:conclusion} concludes the paper.

\section{Related Work}
\label{sec:related-work}

Mathematical reasoning plays a crucial role in artificial intelligence, enabling the comprehension and resolution of intricate mathematical challenges. The incorporation of LLMs in this area has been substantial, thanks to their capability to interpret, process, and produce complex natural language. In artificial intelligence, math problem solving involves utilizing algorithms, computational models, and use of increasingly LLMs to understand, explain, and resolve mathematical challenges. This method encompasses a wide range of topics, from basic arithmetic to advanced mathematics, including areas such as algebra, geometry, statistics, and calculus.
\citep{wei2023chainofthought} boosts the reasoning capacity of LLMs by supplementing the output with a series of intermediate steps leading to the answer. Several approaches have been suggested to enhance the quality of these reasoning paths. For instance, complexity-based CoT \citep{fu2023complexitybased} picks examples with more steps as in-context demonstrations, demonstrating that prompting with additional reasoning steps improves performance. Self-consistency \citep{wang2023selfconsistency} generates multiple reasoning paths and selects the final answer through majority voting. Another set of techniques involves fine-tuning-based methods, which adapt open-source models (like LLaMA) using insights from advanced closed-source LLMs (GPT-4, GPT-3.5-Turbo). \citep{magister-etal-2023-teaching} explore the transfer of reasoning abilities through knowledge distillation. \citep{yuan2024scaling} advocate for the use of rejection sampling fine-tuning (RFT) to enhance mathematical reasoning performance. WizardMath \citep{xu2024wizardlm} introduces a reinforced evol-instruct method for strengthening reasoning abilities through supervised fine-tuning and PPO training \citep{schulman2017proximal}. MAmmoTH \citep{yue2024mammoth} integrates CoT and Program-of-Thought \citep{chen2023program} rationales to teach LLMs how to utilize external tools (such as a Python interpreter) for solving mathematical problems. \citep{wang2023making} propose a constraint alignment loss for fine-tuning LLMs to improve calibration. Going beyond the improvement of mathematical abilities through fine-tuning, LLEMMA \citep{azerbayev2024llemma} introduces the Proof-Pile-2 dataset, which combines mathematical texts and code. By continuously pre-training with Code Llama, the model is equipped to utilize Python interpreters and formal theorem provers, showcasing remarkable performance on the MATH benchmark.



\section{Data}
\label{sec:data}
We followed past works (\citep{ma2024at}, \citep{razeghi2024backtracking}, \citep{aryabumi2024codecodeexploringimpact}) that suggest that source code with text in the pretraining corpus improves the general and mathematical reasoning abilities of generative language models. Thus, we mixed source code related to mathematical problems along with open source mathematical web corpus and clubbed it with our curated lecture notes, and templatised mathematical questions answers in the pretraining dataset.
Our pretraining dataset is a set of selected corpus from various publicly available datasets (AlgebraStack \citep{azerbayev2024llemma}, MathPile Commercial \citep{wang2023mathpile}, AutoMathText \citep{zhang2024automathtext}, and Chain-of-Thought (CoT) templatised StackOverflow math, physics, statistics question answers \citep{stackmathqa2024} and in-house collection of mathematical lecture notes in \LaTeX. AutoMathText from AutoDS \citep{zhang2024automathtext} is a comprehensive and meticulously curated dataset that contains approximately 200 GB of mathematical texts. It is compiled from a variety of sources, including websites, arXiv, and GitHub (OpenWebMath, RedPajama, Algebraic Stack). This extensive dataset has been autonomously selected as zero-shot verifier and labeled by the advanced open-source language model, Qwen-72B. Each item in the dataset is assigned a score, lm\_q1q2\_score, ranging from [0, 1], which indicates its relevance, quality, and educational value in the field of mathematical intelligence.

For our pretraining data, we select only the web corpus from AutoMathText where the  lm\_q1q2\_score $\geq 0.6$. We selected textbooks, proofofwiki, wikipedia, and stackexchange subsets from MathPile Commercial dataset. We selected the source code from AlgebraStack. Finally, the pretraining corpus is composed of mathematical text from web, source code related to mathematical reasoning, one million question-answers pairs from StackOverflow, and in-house math lectures in \LaTeX. Table \ref{tab:data} shows the number of words in our pretraining corpus. We used the following CoT template for templatising the StackOverflow and StackExchange questions answers as part of our pretraining corpus.

''Below is an instruction that describes a task.
Write a response that appropriately completes the request. 
\#\#\# Q:\{question\}
\#\#\# A: Let's think step by step. The answer is: \{answer\}''

\begin{table}[t]
\centering
\begin{tabular}{lr}
\toprule
\textbf{Pre-training Corpus} & \textbf{\# Words} \\ \midrule
Web Corpus from AutoMathText         & 1,245,273,066                         \\
Math Pile Commercial                   & 854,692,279                           \\
Math Code (AlgebraStack)     & 678,729,775                           \\
StackMathQA                  & 297,955,905                           \\
Lecture Notes (ours)         & 2,672,457,786                         \\ \midrule
\bf Total                        & \bf 5,749,108,811   \\                     
\bottomrule
\end{tabular}
\caption{Pretraining Corpus}
\label{tab:data}
\vspace{-2mm}
\end{table}

\subsection{Data Contamination Removal}

Following \citep{kocetkov2023the}, we removed duplicates and near-duplicates from the training data using \citep{chenghao_mou_2023_8364980}, with default parameters. 
Following \citep{guo2024deepseekcoderlargelanguagemodel}, we ran the data decontamination check in order to remove data contamination in the pretraining corpus from the various benchmark evaluations that we performed to test the performance of our math SLM. The filtering criterion is as follows: any text segment containing a 8-gram string that matches exactly with any sub-string from the evaluation benchmarks is removed from our pretraining corpus. For benchmark texts that are shorter than 8 grams but have at least 2 grams, we employ exact matching to filter out contaminated examples. This decontamination process leads us to remove around 170,346,325
words. Finally, the pretraining corpus
has 5,578,762,486 words in the corpus.

\section{Model Design}
\label{sec:model}
The model architecture of Paramanu-Ganita is based on Transformer decoder-only architecture. It uses RMSNorm \citep{10.5555/3454287.3455397} as pre-normalizaion layer with norm\_epsilon = 1e-5, and SwiGLU \citep{shazeer2020glu} activation function in the feed-forward dense layers.
The model, Paramanu-Ganita, uses multi-head attention (MHA). The hidden dimension is 1024 with 16 layers, n\_k\_v\_heads=16, and 16 attention heads with feedforward layer hidden dimension of 2752.
Following \citep{JMLR:v24:22-1144}, we remove all biases from dense layers to improve the training stability of Paramanu-Ganita.

\paragraph{Tokenization} We trained two separate Byte-Pair encoding (BPE) \citep{sennrich-etal-2016-neural} tokenizers using Sentencepiece \citep{kudo-richardson-2018-sentencepiece} module on the pretraining data from scratch to develop mathematical domain specialised tokenizer to learn the intricacies of mathematical terminology. One BPE tokenizer is trained on AlgebraStack (mathematical source code corpus) and another BPE tokenizer is trained on the mathematical text, lecture notes, and StackOverflow question answers. During pre-tokenization, NFC normalization was performed on the processed data, digits are split into individual tokens and fall back unknown UTF-8 characters to byte granularity for improving the arithmetic learning ability of the pretrained model. We treat our data as a sequence of bytes rather than Unicode characters, and we include each of the 256 bytes as tokens. We then merged the both mathematical tokenizer and code tokenizer by intersection by removing the duplicate tokens to develop our final tokenizer specialised in mathematics and code, compact, optimized, and effective. Tokenizer has special tokens like ``\textlangle Q:\textrangle'',  ``\textlangle A:\textrangle'', ``\textlangle tex\textrangle'', ``\textlangle/tex\textrangle'', ``\textlangle python\textrangle'', ``\textlangle/python\textrangle'', ``\textlangle c\textrangle'', ``\textlangle/c\textrangle'', ``\textlangle matlab\textrangle'', ``\textlangle/matlab\textrangle'' ``\textlangle haskell \textrangle'', ``\textlangle/haskell\textrangle''. 

\section{Training of SLM}
\label{sec:training}
\subsection{Pre-training}
We have pretrained our math SLM, Paramanu-Ganita, from scratch at a context size of 4096 on our curated corpus. However, we have excluded training of our model on ArXiv math papers as we believe that to learn basic mathematical concepts, and acquire mathematical logical reasoning, ArXiv math papers are not required as they generally meant to serve beyond high school level mathematics. We started with simple strategy to use a part of our curated corpus which generally covers various mathematical and logical concepts till secondary school education in general. We performed mix pretraining combining both mathematical plain text, source code of programming languages, and templatised mathematical question answers pairs in the pretraining phase. For pretraining Paramanu-Ganita (4096 context size), we performed 95\%-5\% data split for pretraining. The perplexity of our model is 4.349 while the MFU is 40.392.

\begin{figure*}[t]
    \centering
    \vspace*{0mm}
    \includegraphics[width=0.6\linewidth]{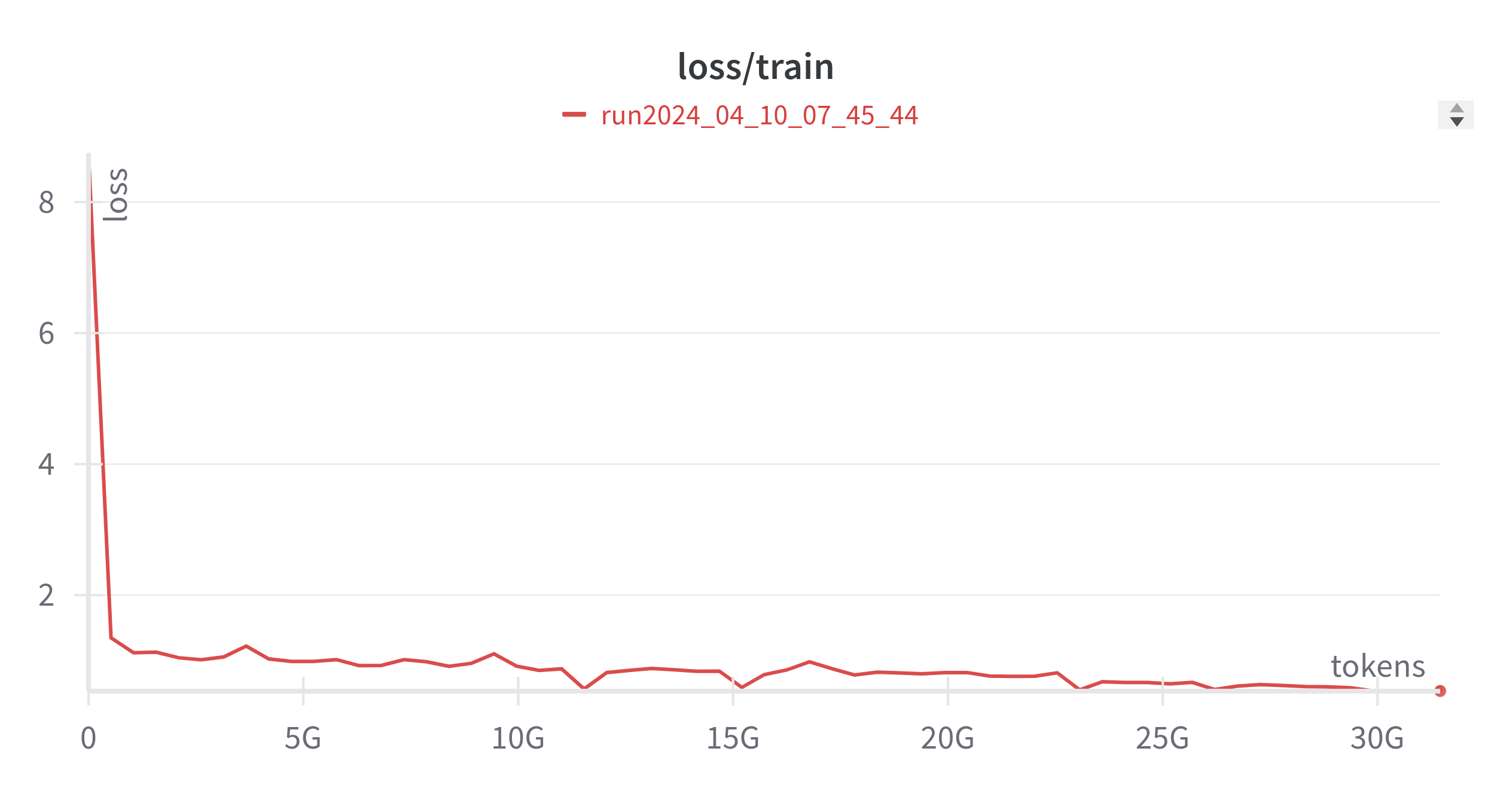}
    \vspace*{-2mm}
    \caption{Training loss against number of tokens in billion. (G = billion)}
    \label{fig:train-loss}
    \vspace*{0mm}
\end{figure*}

We performed hyperparameter tuning on 15M models to find the optimal vocabulary size, learning rate, learning rate scheduler, and warm-up ratio. We used a batch size of 8, gradient accumulation steps of 8, and the maximum sequence length set to 4096, i.e., 262,144 tokens per iteration. We used the concept of $\mu$ transfer, and transferred the learned hyperparameters to our bigger model for 208M Paramanu-Ganita from 15M model. Following \citep{hoffmann2022training}, we set $lr$ decay steps to $max\_steps$ and the minimum $lr$ is set nearly to 0.1$\cdot lr$. The $lr$ schedule starts with a linear warm-up from 0 to the maximum $lr$ at 1000 steps, followed by a cosine decay to the minimum $lr$ until the $max\_steps$ = 120,000 end of an epoch of training. We used the following equation for $lr$ decay ratio.
\begin{align*}
	lr_{decay\_ratio} = \frac{t - warmup_{steps}}{lr_{decay\_steps} - warmup_{steps}}
\end{align*}
where $t$ is the current training step.
We set maximum learning rate ($lr$) to 3e-3 (max), weight decay to 1e-1. To further speedup training, we used BF16 mixed precision training. For our experiments and modeling, we implemented our code using Pytorch 2.0, in-house optmized CUDA kernels and used \texttt{torch.compile} feature for every model. We can see from Figure~\ref{fig:train-loss} how the loss is converging with incremental training steps and pretraining tokens, confirming a good pretraining with minor loss spikes. Paramanu-Ganita 208M is pretrained on around a total of 31.5 billion tokens.

\subsection{Chain-of-Thought Instruction Fine-tuning}

We performed Chain-of-Thought instruction fine-tuning on the MetaMathQA \citep{yu2024metamath} instructions dataset, i.e, instead of regular instruction fine-tuning, we prepend the response of the MetaMathQA dataset by a prompt ``Let's think step by step'', and then used the prepended instruction, and response pair for instruction fine-tuning. We fine-tuned for 2 epochs due to limited computational resources. We used cosine learning rate scheduler with ($lr$) set to 2e-5 with gradient clipping of 1.0, warmup ratio of 0.05 and no weight decay.
However, we believe our instruction-tuned Paramanu-Ganita would have performed better in benchmark evaluation if it was further fine-tuned for another 2-3 epochs.  
We used the following training prompt for MetaMathQA for our model. 

''Below is an instruction that describes a task.
Write a response that appropriately completes the request. 
\#\#\# Q:\{query\}
\#\#\# A: Let's think step by step. \{response\}''
 
\section{Evaluation}
\label{sec:eval}
In this section, we present results of our model Paramanu-Ganita against different LLMs, both general and code LLMs as well as math-specialized, on various benchmarks. We evaluated across variety of grade level of difficulty benchmarks including both discriminative multiple-choice math benchmarks across grade school level, high school level, college level, competitive exams level of SAT, GRE, GMAT, and math competition level questions. We also tested our model on a logical reasoning benchmark (LogiQA).


\subsection{RQ1: GSM8K and MATH benchmark datasets}
We evaluate the model's ability to solve mathematics problems using chain of thought reasoning. Our evaluations include GSM8K \citep{cobbe2021training} and MATH \citep{hendrycks2021measuringmath}, which are the standard benchmarks for evaluating quantitative reasoning in language models. GSM8K includes 8,500 high-quality grade school math problems created by human writers. These problems generally consist of 2 to 8 steps to solve and mainly involve a series of basic arithmetic calculations to arrive at the final answer. The MATH dataset consists of 12,500 problems taken from high school math competitions. Each problem includes a step-by-step solution, allowing models to learn how to generate answer derivations and explanations.
We used the following evaluation prompt for GSM8K test set for our math model. 

''Below is an instruction that describes a task.
Write a response that appropriately completes the request. 
\#\#\# Q:\{question\}
\#\#\# A: Let's think step by step. The answer is: ''

We used the vLLM \citep{10.1145/3600006.3613165} inference engine and used the code from \citep{yu2024metamath} for evaluation on GSM8K and MATH benchmarks. The following stop tokens were used while decoding from the model during evaluation.  stop=[`Question:', `Question', `USER:', `USER', `ASSISTANT:', `ASSISTANT', `Instruction:', `Instruction', `Response:', `Response'] We set the vLLM inference engine parameters: best\_of=8, presence\_penalty=0.0, frequency\_penalty=0.0, repetition\_penalty=1.0, temperature=0.1, top\_p=1, top\_k=-1, min\_p=0.0,  length\_penalty=1.0, max\_tokens=1024 while decoding from Paramanu-Ganita for evaluation on GSM8K and Math test benchmarks. Answer extraction differs from the method used by \citep{wei2023chainofthought} who rely on complex string rules to derive the final answer. In contrast, we follow the approach of WizardMath \citep{luo2023wizardmathempoweringmathematicalreasoning} by only extracting the string that follows ``The answer is:'' as the final answer. To train the model on this extraction technique, we append ``The answer is: {gold answer}'' to the end of the answers in the MetaMathQA dataset, replacing the gold answer with the corresponding answer for each question.

We report accuracy of Paramanu-Ganita and other models in Table~\ref{tab:gsm8k-math}. 
The scores of these models are reproduced as-is from their respective publications.
 

\begin{table*}[t]
\centering
\resizebox{\columnwidth}{!}{\begin{tabular}{ l rrr}
\toprule
\textbf{Model} & \textbf{Parameters}      & \textbf{GSM8K} & \textbf{MATH}\\ \midrule
LLaMa-1 & 7B           & 11.00     &  2.90          \\ 
LLaMa-1 & 13B	& 17.80 &	3.90 \\ 
LLaMa-1 & 33B          & 35.60     &   3.90          \\ 
LLaMa-2 & 7B           & 14.60     &  2.50           \\ 
LLaMa-2 & 13B          & 28.70     &   3.90          \\ 
Code Llama & 7B           & 10.50  &   13.00            \\
Code Llama & 13B & 36.10 & 16.40 \\ 
Code Llama & 34B           & 29.60  & 12.20               \\
Falcon & 40B           & 19.60  & 2.50                \\ 
Falcon & 7B            & 6.80   & 2.30                \\ 
MPT & 30B              & 15.20  & 3.10               \\ 
MPT & 7B               & 6.80   &   3.00              \\ 
GPT-J & 6B               & 34.90  &  -               \\
Vicuna & 13B           & 27.60   &   -            \\ 
PaLM & 8B              & 4.10    &  1.50             \\ 
PaLM & 62B             & 33.00   &   4.40            \\ 
Minerva & 8B            & 16.20  &   14.10             \\ 
Minerva & 62B           & 52.40  &    27.60            \\ 
Minerva & 540B         & 58.80   &  33.60             \\ 
MAmooTH-CoT & 7B & 50.50 & 10.40 \\ 
WizardMath & 7B & 54.90 & 10.70 \\ 
MetaMath & 7B & 66.50 & 19.80 \\ 
LLEMMA & 7B           & 36.40    &   18.00           \\ 
LLEMMA & 34B           & 51.50   &    25.00           \\ \midrule
Paramanu-Ganita & 208M & 36.77   & 10.34     \\ \bottomrule
\end{tabular}}
\caption{Evaluation of LLMs on GSM8K test set. PaLM \citep{JMLR:v24:22-1144}, LLaMa-1 \citep{touvron2023llama}, LLaMa-2 \citep{touvron2023llama2}, Falcon \citep{almazrouei2023falcon}, Code LlaMa \citep{rozière2024code}, MPT \citep{databricksIntroducingMPT7B}, Vicuna \citep{vicuna2023}, Minerva \citep{lewkowycz2022solving}, MAmooTH-CoT \citep{yue2024mammoth}, MetaMath \citep{yu2024metamath}, WizardMath \citep{luo2023wizardmathempoweringmathematicalreasoning}, LLEMMA \citep{azerbayev2024llemma} scores are quoted from respective author papers.}
\label{tab:gsm8k-math}
\vspace*{0mm}
\end{table*}

\subsection{RQ1: Multiple-choice Math QA benchmark datasets}
We evaluate our model and compare with LLMs including general LLMs, math-specialized LLMs like LLEMMA, and code LLMs like CodeLlama on various multiple choice math question answers using lm-eval-harness \citep{lintang_sutawika_2024_10600400} at zero-shot greedy decoding setting. We considered high school and college math MCQ question answers from MMLU \citep{hendrycks2021measuring}, AGIEVAL-AQuA-RAT (GRE, GMAT 254 multiple-choice math questions taken from AQuA-RAT \citep{ling-etal-2017-program}) \citep{zhong-etal-2024-agieval}, and AGIEVAL-SAT-Math (SAT 220 multiple-choice math questions). Table~\ref{tab:mcq-math} compares Paramanu-Ganita with various LLMs. LogiQA \citep{10.5555/3491440.3491941} is a dataset created from various logical reasoning questions gathered from China's National Civil Servants Examination. Notably, LogiQA features bilingual questions in both English and Chinese, with the English version being a translation of the original Chinese text. We only considered the English version for evaluation.

\begin{table*}[t]
\centering
\resizebox{\linewidth}{!}{
\begin{tabular}{l r r r r r}
\toprule
\textbf{Models}          & \textbf{LogiQA} & \textbf{MMLU-math-} & \textbf{MMLU-math-} & \textbf{AGIEVAL-} & \textbf{AGIEVAL-} \\
& & \textbf{high-school} & \textbf{college} & \textbf{AQuA-RAT} & \textbf{SAT-Math} \\
\midrule
Llama-2 7B      & 30.41                                & 25.55                                       & 30.00                                   & 25.59                               & 24.54                                \\
CodeLlama-7B    & 30.72                                & 24.81                                       & 30.00                                   & 22.83                               & 29.09                                \\
OLMo 1B         & 26.81                                & 30.37                                       & 27.00                                   & 23.62                               & 21.81                                \\
LLEMMA 7B       & 29.95                                & 32.22                                       & 32.00                                   & 23.22                               & 32.72                                \\
Falcon 7B       & 26.88                                & 21.11                                       & 21.00                                   & 22.04                               & 28.63                                \\
\midrule
Paramanu-Ganita 208M & 30.57                                & 31.11                                       & 29.00                                   & 26.77                               & 25.00        \\ \bottomrule                          
\end{tabular}}
\caption{Zero-shot evaluation of Paramanu-Ganita 208M and LLMs. All benchmark reports Accuracy except LogiQA, which reports Normalized Accuracy. We present the best score across our model checkpoints for Paramanu-Ganita. B=billion, M=million.}
\label{tab:mcq-math}
\vspace*{0mm}
\end{table*}

\vspace{-2mm}
\section{Results and Analysis}
\label{sec:analysis}

From Table ~\ref{tab:gsm8k-math} on GSM8K benchmark, Paramanu-Ganita, despite being 35 times smaller than the 7B family of LLMs, outperformed LLaMa-1 7B by 25.8\% points, LLaMa-2 7B by 22.17\% points, Falcon 7B by 29.97\% points, PaLM 8B by 32.67\% points, Minerva 8B by 20.6\% points, and LLEMMA-7B respectively. Paramanu-Ganita also outperformed PaLM 62B by 3.8\% points despite being smaller by 305 times, Falcon 40B by 17.2\% points (smaller by 192 times), LLaMa-1 33B by 1.2\% points (smaller by 158 times), and Vicuna 13B by 9.2\% (smaller by 64 times).
This is a significant achievement since smaller models are preferred due to cost and environmental sustainability.
Only the 3 giant LLMs, namely, LLEMMA 34B, Minerva 62B, Minerva 540B, performed better than Paramanu-Ganita on the GSM8K benchmark.

On the MATH benchmark as shown in the Table~\ref{tab:gsm8k-math}, Paramanu-Ganita outperformed LLaMa-1 7B by 7.44\%, Llama-1 13B by 6.44\% points, Llama-2 7B by 7.84\% points, Llama-2 13B by 6.44\% points, Falcon 7B by 8.04\% points, Falcon 40B by 7.84\% points, MPT 30B by 7.24\% points, MPT 30B by 7.24\% points, PaLM 8B by 8.84\% points, and PaLM 62B by 5.94\% points respectively. GPT-J and Vicuna did not report numbers for the MATH benchmark.

As shown in the Table~\ref{tab:mcq-math} on LogiQA \citep{10.5555/3491440.3491941} benchmark, Paramanu-Ganita outperformed Llama-2 7B, OLMo 1B by 3.76\% points, LLEMMA 7B, Falcon 7B by 3.69\% points.

On mathematical high school questions (MMLU-math-high-school) benchmark as shown in the Table~\ref{tab:mcq-math}, Paramanu-Ganita outperformed Llama-2 7B by 5.56\% points, CodeLlama 7B by 6.3\% points, OLMo 1B, and Falcon 7B by 10\% points but LLEMMA 7B outperformed Ganita by 1\% point despite being 34 times larger in size.

On college level math questions (MMLU-math-college) benchmark as shown in the Table~\ref{tab:mcq-math}, Paramanu-Ganita 208M outperformed Falcon 7B by 8\% points, OLMo 1B by 2\% points, whereas both Llama-2 7B and CodeLlama 7B outperformed Paramanu-Ganita just by 1\% point despite being 34 times larger.

On GRE-GMAT level quantitative questions (AGIEVAL-AQuA-RAT) benchmark as shown in Table~\ref{tab:mcq-math}, Paramanu-Ganita outperformed all the LLMs under comparison, i.e., Falcon 7B by 4.73\% points, LLEMMA 7B by 3.55\% points, OLMo 1B by 3.15\% points, CodeLlama 7B by 3.94\% points, and Llama-2 7B by 1.18\% point respectively.

At SAT level math questions (AGIEVAL-SAR-Math) benchmark as listed in Table~\ref{tab:mcq-math}, Paramanu-Ganita outperformed Llama-2 7B, and OLMo 1B while lagging behind LLEMMA 7B by 7.72\% points.

Despite being 35 times smaller in size, the performance of our model, Paramanu-Ganita, is comparable with the other LLMs. 
We believe the domain specific pretraining from scratch using high quality mathematical corpus of lecture notes, source code, web scrapped mathematical text and our Chain-of-Thought (CoT) templated formatted StackOverflow math, physics, statistics question answers along with our novel merged math and code specialized BPE tokenizer, and CoT instruction fine-tuning are the most probable causes to amplify the performance of strong mathematical reasoning in a small generative language model of 208 million parameters compared to LLMs which are pretrained on all kinds of data whereas we focused only on mathematical and source code related to mathematics, mathematical question answers in COT template in our pretraining corpus. 

\section{Conclusions and Future Work}
\label{sec:conclusion}
In this paper, we explore the performance of small math language model compared to both generalist and math specialised LLMs. Instead of continual pretraining of LLMs and then applying various fine-tuning approaches to improve the reasoning of LLMs, we introduce an exclusive decoder-only math SLM, Paramanu-Ganita 208M, pretrained from scratch on a diverse mathematical corpus, including web text, textbooks,  \LaTeX{} lecture notes, programming code, and CoT templatised mathematical question-answer pairs. We fine-tuned Paramanu-Ganita using CoT instructions on the MetaMathQA dataset. Our model was evaluated across various benchmarks at the grade school, high school, college, and competitive exam levels (SAT, GRE, GMAT), and outperformed generalist LLMs and math-specific models, such as Minerva 8B and LLEMMA 7B, in accuracy on GSM8K and MATH benchmarks. Paramanu-Ganita also achieved competitive results on logical reasoning (LogiQA), (SAT, GRE, GMAT) math subsets of AGIEVAL, and math questions from MMLU, outperforming or matching larger 7B LLMs.

Our extensive evaluation of Paramanu-Ganita and other LLMs on mathematical and logical reasoning tasks leads to the conclusion that a small, domain-specific language model, when trained from scratch with a specialized tokenizer, is a more cost-effective and environmentally friendly approach. With only 170 A100 training hours, Paramanu-Ganita offers a significant reduction in training costs, i.e., 135 times less than the continual pretraining of Llama for mathematical reasoning \citep{azerbayev2024llemma} without sacrificing performance on key math benchmarks. This work challenges the ``bigger is better'' presumption, demonstrating the potential and effectiveness of creating small domain specific language models from scratch rather than relying on existing LLMs.

For future work, we plan to expand the pretraining corpus with ArXiv math papers and explore reinforcement learning alignment (e.g., DPO \citep{rafailov2024directpreferenceoptimizationlanguage}) to further enhance our model's performance.

\section{Ethics Statement}

We have used results of the other models from their respective publications. We have trained and evaluated our model on a single GPU. Thus, we do not see any issues of ethics for this work.

\bibliographystyle{acl_natbib}
\bibliography{anthology,custom}


\section*{Acknowledgements}
The first author wants to dedicate his work to his
beloved parents, Rita Niyogi and Malay Niyogi for
their outstanding support throughout his journey.

\appendix

\section{Background}

\subsection{Language Modeling}

This objective of the language modeling can be formally described as maximizing the probability of a sequence of tokens $w_1, w_2, \dots, w_{N}$
\begin{align*}
	P(w_{1}, w_{2}, \ldots, w_{n}) = \prod_{i=1}^{n} P(w_i \,|\, w_1, w_2, \ldots, w_{i-1})
\end{align*}
where $p(w_{t}|w_{0}, \dots w_{t-1})$ is the probability of token $w_{t}$ given the sequence of previous tokens $w_{0}, \dots, w_{t-1}$.

The performance of a language model is generally evaluated using the total \emph{cross-entropy loss}, i.e, the negative log-likelihood of the observed data under the model under consideration, which for a given
dataset is defined as: 
\begin{align*}
	   Loss = -\frac{1}{N} \sum_{i=1}^{n} \log( P(w_i \,|\, w_1, w_2, \ldots, w_{i-1}) )
\end{align*}
Lower the loss better is the model; however, just computing the loss may not be intuitive. Therefore, \emph{Perplexity} is a metric to evaluate the performance of a given language model which is the exponent of the average loss.
\begin{align*}
	Perplexity = \exp\left( Loss \right)
\end{align*}

\subsection{Rotary Position Embedding (RoPE)}

Transformer-based models rely on positional embeddings to encode position and relative location information of words in a text. 
\emph{Rotary Position Embedding (RoPE)} is a position encoding technique proposed by \citep{black-etal-2022-gpt}.
Instead of adding positional embeddings or relative positional embeddings to token embeddings, RoPE rotates the token embedding by a fixed factor ($\theta$) in the higher-dimensional space to encode relative positional embeddings. In other words, RoPE encodes the absolute positions with a rotation matrix and meanwhile incorporates the explicit relative position dependency in self-attention formulation. The intuition behind RoPE is that we can represent the token embeddings as complex numbers and their positions as pure rotations that we apply to them. If we shift both the query and key by the same amount, changing absolute position but not relative position, this will lead both representations to be additionally rotated in the same manner. Thus, the angle between them will remain unchanged and, thus, the dot product will also remain unchanged. By exploiting the nature of rotations, the dot product used in self-attention will have the property for preserving relative positional information while discarding absolute position.

\subsection{Model FLOPs Utilization (MFU)} 
Model FLOPs Utilization (MFU) \citep{JMLR:v24:22-1144} estimate is the ratio of the observed throughput (tokens-per-second) relative to the theoretical maximum throughput of a system at peak FLOPs. Model flops utilization (MFU) estimate the number of flops (floating point operations) done per iteration. It quantifies how efficiently the GPUs are utilized in model training.

\subsection{Maximal Update Parameterization}
As the size of large language models (LLMs) and the scale of the dataset used in pretraining are expensively large, it is not feasible to perform hyperparameter tuning in LLMs. \citet{yang2021tuning} used a technique called maximal update parameterization ($\mu P$) to  transfer the hyperparameters learnt from tuning of a small model to a larger model and found that the optimal hyperparameter values become stable across neural network sizes when the models have been parameterized using ($\mu P$).

\subsection{Root Mean Square Normalization (RMSNorm)}

To improve the training stability, some LLMs (Chinchilla \citep{NEURIPS2022_c1e2faff}, LLaMa \citep{touvron2023llama}) have normalized the input of each transformer sub-layer, instead of normalizing the output using RMSNorm normalizing function as introduced by \citep{10.5555/3454287.3455397}. 
\emph{RMSNorm} normalizes the activations based on their root mean square (RMS) value instead of normalizing the inputs based on their mean and variance.

RMSNorm accelerates the training and inference with similar performance in these large models. It is reported that replacing LayerNorm \citep{ba2016layer} with RMSNorm can achieve comparable performance and improve training and inference time by 7-64\%.
\citet{narang2021transformer} showed that RMSNorm improves the pre-training speed by 5\% compared with the LayerNorm baseline.


\end{document}